\def\myauthor{Jose Cerqueira Fernandes and Benjamin Kenwright}

\def\myorganisation{Heriot-Watt University}

\def\mytitle{Identifying and Extracting Football Features \\ from Real-World Media Sources using Only Synthetic Training Data}

\def\mykeywords{graphics, machine learning, football, analytics, machine learning, generating, training, procedural}

\def\myabstract{
Real-world images used for training machine learning algorithms are often unstructured and inconsistent.
The process of analysing and tagging these images can be costly and error prone (also availability, gaps and legal conundrums).
However, as we demonstrate in this article, the potential to generate accurate graphical images that are indistinguishable from real-world sources has a multitude of benefits in machine learning paradigms.
One such example of this is football data from broadcast services (television and other streaming media sources).
The football games are usually recorded from multiple sources (cameras and phones) and resolutions, not to mention, occlusion of visual details and other artefacts (like blurring, weathering and lighting conditions) which make it difficult to accurately identify features.
We demonstrate an approach which is able to overcome these limitations using generated tagged and structured images.
The generated images are able to simulate a variety views and conditions (including noise and blurring) which may only occur sporadically in real-world data and make it difficult for machine learning algorithm to `cope' with these unforeseen problems in real-data.
This approach enables us to rapidly train and prepare a robust solution that accurately extracts features (e.g., spacial locations, markers on the pitch, player positions, ball location and camera FOV) from real-world football match sources for analytical purposes.

}

\documentclass[8pt,sigconf]{acmart}

\usepackage{multicol}
\hypersetup{
    colorlinks = true, 
    linkcolor = black,
    anchorcolor = red,
    citecolor = blue, 
    filecolor = black, 
}

\usepackage{enumitem}
\setlist[itemize]{noitemsep,leftmargin=*}
\setlist[enumerate]{noitemsep,leftmargin=*}

\usepackage{titlesec}
\titleformat{\paragraph}[runin]
{\normalfont\normalsize\bfseries}{\theparagraph}{}{}
\titlespacing*{\paragraph}{0pt}{3pt}{8pt}

\usepackage{listings}
\usepackage{xcolor}

\definecolor{codegreen}{rgb}{0,0.6,0}
\definecolor{codegray}{rgb}{0.5,0.5,0.5}
\definecolor{codepurple}{rgb}{0.58,0,0.82}
\definecolor{backcolour}{rgb}{0.95,0.95,0.92}
\lstdefinestyle{mystyle}{
    columns=fixed,
	basewidth=0.4em, % ***************** important line ************
    backgroundcolor=\color{backcolour},   
    commentstyle=\color{codegreen},
    keywordstyle=\color{magenta},
    numberstyle=\tiny\color{codegray},
    stringstyle=\color{codepurple},
    basicstyle=\ttfamily\footnotesize,
    breakatwhitespace=false,         
    breaklines=true,                 
    captionpos=b,                    
    keepspaces=true,                                
    showspaces=false,                
    showstringspaces=false,
    showtabs=false,                  
    tabsize=2
}

\usepackage{graphicx}

\graphicspath{{./images/}}

\newcommand{\figuremacroW}[4]{
	\begin{figure}[!htbp] % [t] %[htbp]
		\centering
		\includegraphics[width=#4\columnwidth]{#1}
		\caption[#2]{\textbf{#2} - #3}
		\label{fig:#1}
	\end{figure}
}

\newcommand{\figuremacroF}[4]{
	\begin{figure*}[!htbp]
		\centering
		\includegraphics[width=#4\textwidth]{#1}
		\caption[#2]{\textbf{#2} - #3}
		\label{fig:#1}
	\end{figure*}
}

\setcopyright{rightsretained}
\copyrightyear{2022}
\acmYear{2022}

\acmConference{Article}{August, 2022}{Synthetic Football Training Data}
\acmISBN{ }

\begin{document}

\title{ \mytitle  \vspace{-20pt} }

\author{\myauthor}
\orcid{0000-0002-4889-9110}

\affiliation{%
  \institution{\myorganisation}
}

\renewcommand{\shortauthors}{\myauthor}
\fontsize{9.2}{9.2}\selectfont

\begin{abstract}
\myabstract
\end{abstract}




\raggedbottom
\sloppy

\maketitle

\raggedbottom
\sloppy

\vspace{-5pt}

\section{Introduction}

Manual feature tagging and extraction in machine learning is tedious (if not impractical) and error prone.
We demonstrate that it is possible to perform computer vision feature extraction on raw football game images using a machine learning solution that was trained using only synthetic data.
The football community has long enjoyed the benefits of using machine learning techniques to extract and analyze details (from individual player performance to the overall game statistics \cite{herold2019machine}). Common to use real-world data sources (recordings of live games) as they captures details which are vital for analytics. Identifying and extracting the features for these analytical processes is often difficult and painstaking slow (especially for raw/live football matches). While researchers have tried to improve the methods for tagging and extracting information from football images, little research has been done on using synthetic models for training the system. We show that it is possible to synthesize data with minimal domain gap (difference between the synthetic and real-data), so that models trained on synthetic data generalize to real in-the-wild datasets.

We describe how to combine 3-dimensional graphical models (environments, objects and players) with a rendering framework to create image datasets with unprecedented realism and diversity. We train a machine learning system using the generated dataset to identify feature such as landmark localizations.
We demonstrate that synthetic data can be used to accurately locate key features in real-world data as well as open up new approaches where manual labelling would be impossible.

\figuremacroW
{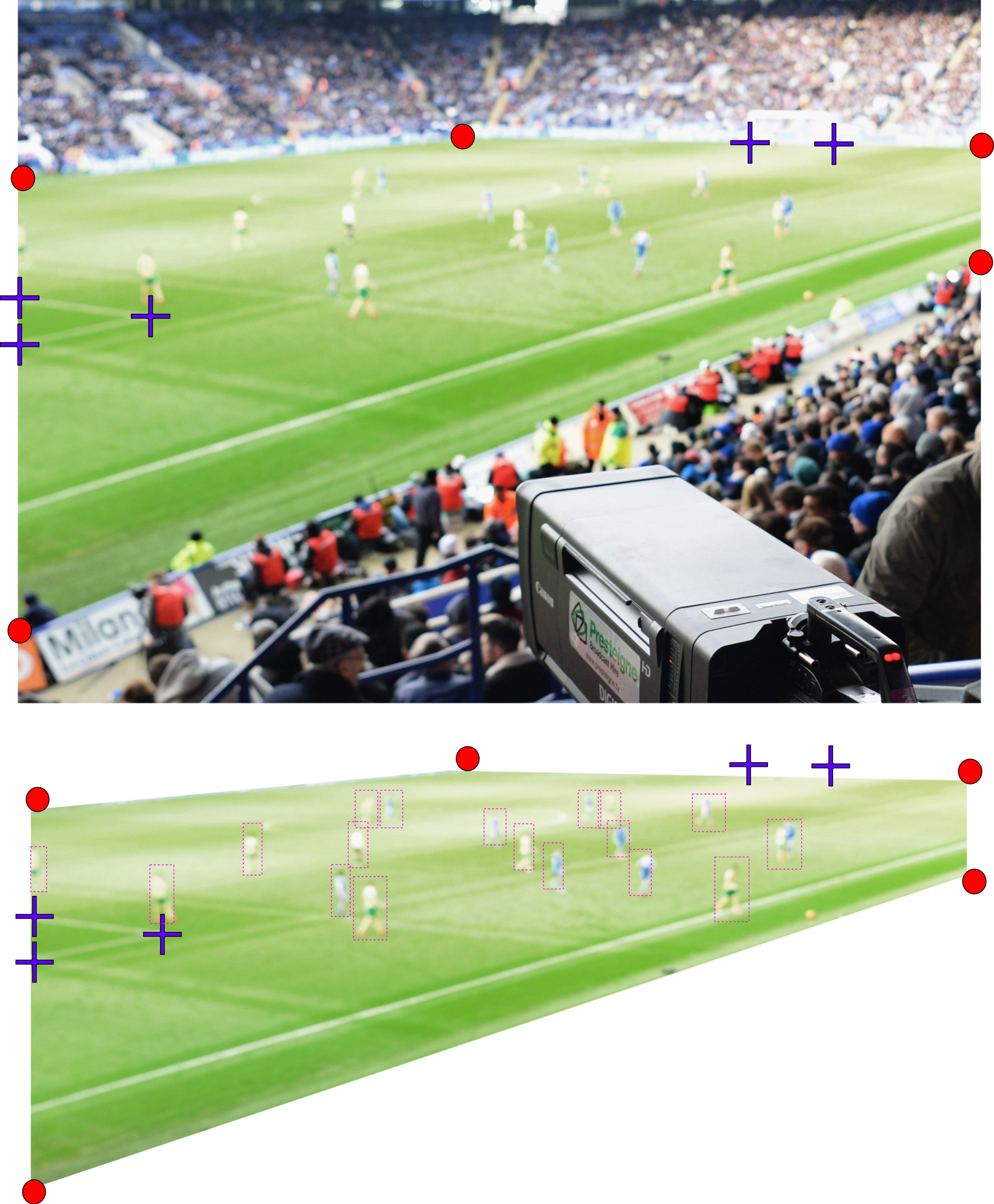}
{Live Image Data}
{Ability to identify and extrapolate key areas from real-world images (football pitch and object markers) in real-time without human intervention (this includes images with poor resolution, blurring, artifacts and occluded details). }
{1.0}

\paragraph{Contributions}
The main contributions of this work are:
(1) novel technique for synthesizing labeled football images for training machine learning algorithms;
(2) demonstration of the capability of synthesizing features of arbitrary geometries and their corresponding labeled images; 
(3) use of the synthesized data for training machine-learning based feature parameter extractors;
and
(4) open source datasets for testing of synthetic training data.

\section{Related Work}

Feature matching machine-learning algorithms require a huge pool of labeled data for proper training, which is often unavailable \cite{konnik2021training}.
To resolve this shortage generated images can be used in place of real-world images.
The concept of using generated graphical images for training machine learning systems is not a new one, and has been used other areas, such as face analysis \cite{wood2021fake} and feature extraction (x-rays images \cite{konnik2021training}).

The most common approach for automatically identifying features in computer-vision (CV) techniques is to segment the images into distinct partitions \cite{blaschke2006object,kaur2014various,zaitoun2015survey} which hopefully provide meaning for extracting geometric information.
 
Currently all of the publications on machine learning and football feature extraction have used real-world data as the source for the training models \cite{herold2019machine}. Most of these data extraction projects have focused on understanding the game mechanics (predicting and understanding the game) \cite{rommers2020machine,herold2019machine}.

\paragraph{Our work:} As far as we are aware no other work has attempted to fully generate a synthetic training set using computer graphics for use on real-world football game feature extraction. 

\figuremacroW
{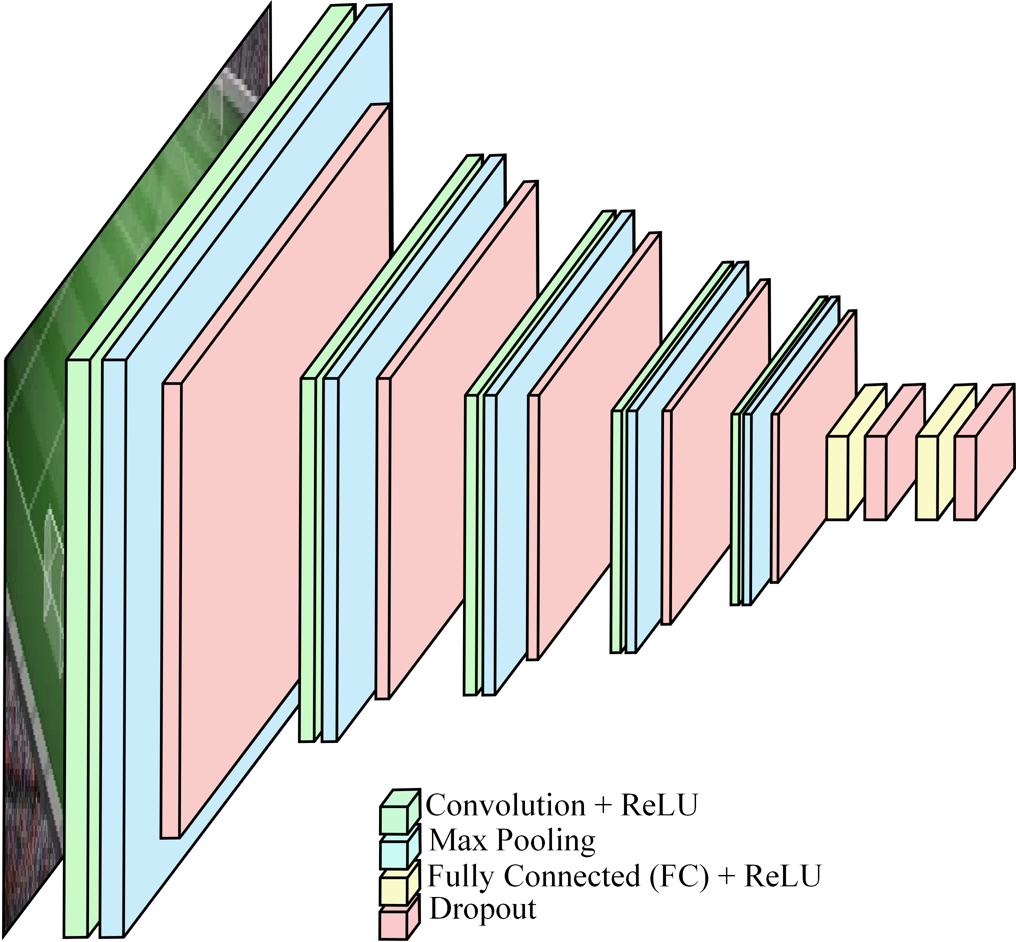}
{Network Architecture}
{Keypoint extraction using a CNN.  The figure summarizes the blocks of the convolution layers of the approach presented in this paper. For example, given a input of 256x256 pixels we extracted 52 dimensional features (26 x-y points identifying key locations on the pitch - as shown in Figure \ref{fig:pitch-markers}). }
{0.7}

\section{Method}

\figuremacroF
{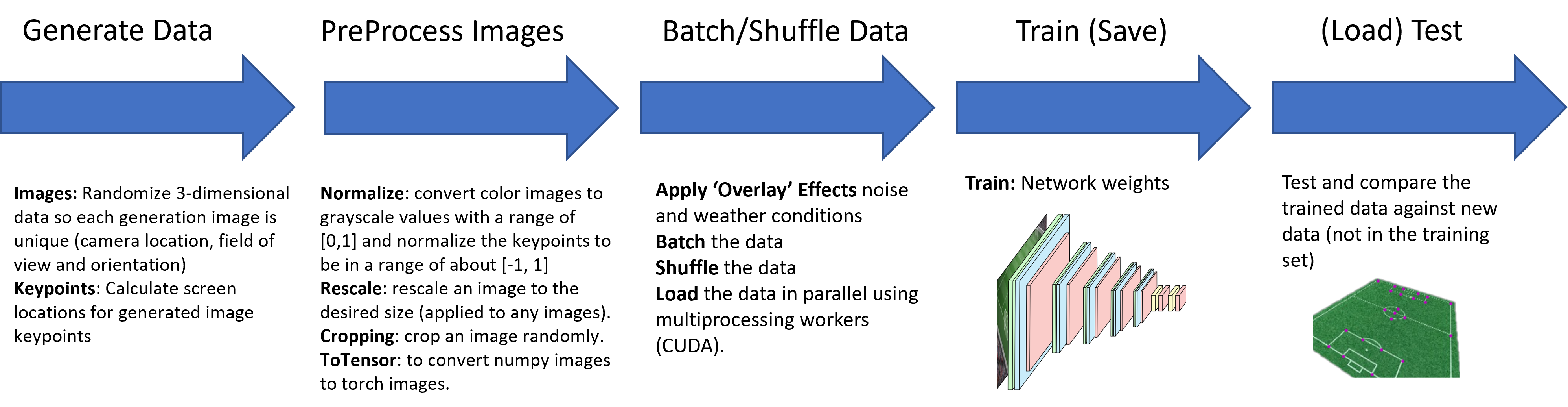}
{Stages}
{Process of creating the dataset through to training and evaluation.}
{1.0}

\paragraph{Training methodology}
We rendered multiple image datasets with a focus on specific features.

During training, we perform data augmentation with overlay images. These augmentations were especially important for synthetic images which would otherwise be free of imperfections (noise and other artifacts). While some of these could be done at render time, we performed them at training time in order to randomly apply different augmentations to the same training images.
We implemented neural networks with PyTorch and trained them with the Adam optimizer.
Since the football pitch sits on a flat ground plan, a top down layout was used to define the pitch keypoint markers (shown in Figure \ref{fig:pitch-markers}).

\figuremacroW
{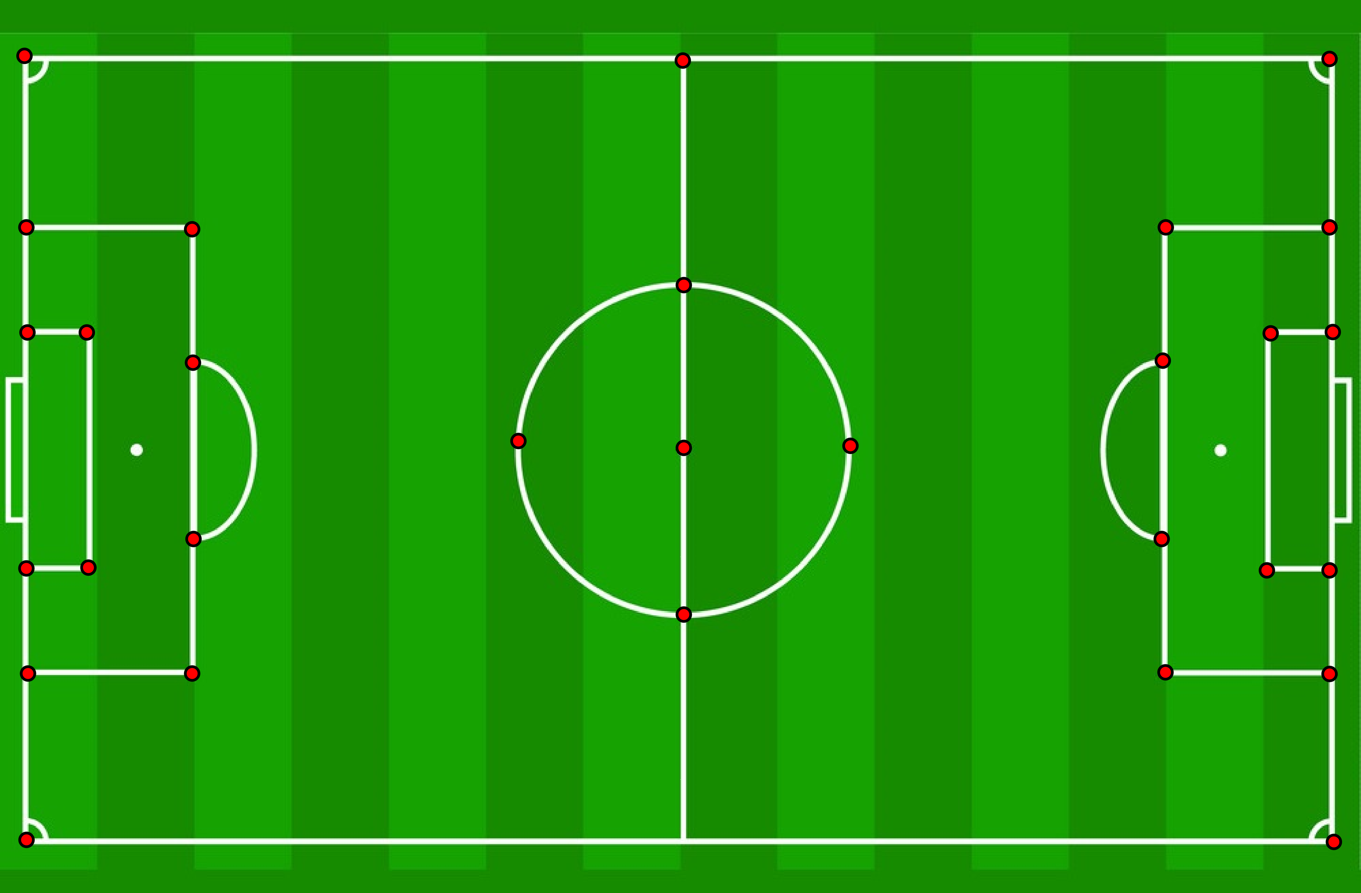}
{Pitch Markers (KeyPoints)}
{The field markers are essential as they provide vital information (e.g., players location during the game). As shown in Figure \ref{fig:lighting}, the view of the game might only show a limited region (field markers help the machine learning algorithm identify the environmental positions).  Consists of 7 horizontal lines, 10 vertical lines and 3 arcs. }
{0.8}

\paragraph{Training Data and Test Cases}
The experiments comprised of different datasets with 3000 images for training and 100 images for testing.  The different datasets evaluated the value and impact of different synthetic features (for example, the value of adding or removing lighting from a synthetic scene also the importance of a realistic surroundings vs blank or randomly placed graphics).

\begin{itemize}
\item Flat pitch (blank background, no lighting or players) as shown in Figure \ref{fig:pitch-markers}.
\item Pitch with random pictures in the background (add noise and feature distractions to the training)
\item Pitch with random lighting conditions
\item Pitch players randomly placed on pitch
\item Pitch with stadium model and random lighting conditions
\item Samples with visual artifacts added to simulate various conditions (e.g., weather or image corruption), using overlay textures (to noise, corruption, blurring).
\end{itemize}

\paragraph{Landmark (Pitch) localization}

Landmark localization finds the position for the pitch (2d locations on screen and their corresponding 3-dimensional locations relative to the camera).
We evaluate our approach using synthetic data (both controlled test cases and generate ones to evaluate a diverse range of problems around extracting features from football images).

The trained network used a mean squared error loss to directly predict. 
We use the provided feature points to extract and identify regions of interest from the image (e.g., pixel regions-of-interest from each image). 
As shown in Figure \ref{fig:pitch-markers}, the ground markers provide essential details (e.g., location of players during the game).  Importantly, multiple image sources could be used to extrapolate the complete game data, which means, different images sources may only show specific regions of the game at specific times (including different viewing angles, lighting, image artifacts and so on).  Most of the time from live media footage of football games, it is rare to see all the players and all the information in high resolution at the same time in a single image.

\paragraph{2-dimensional data to 3-dimensional data}
The football pitch is flat and conforms to a international standard, so the 2d pitch key-point data can be used as a point of reference for reconstructing a relative 3-dimensional layout of the game.  This data feeds through to player position data, using the key-point data lets us calculate the location of the player in real-world coordinates.

% -----------------------\\

\figuremacroW
{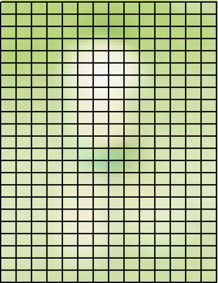}
{Resolution}
{Often features are low-resolution (e.g., players); usually only taking up a few pixels in the bigger picture. For instance, the image shows a zoomed in view of one of the players from Figure \ref{fig:teaser}.  Note the player is more identifiable when the image is considered in context of the surrounding's (zoomed out versus in isolation).}
{0.2}

Our synthetic football images are realistic, diverse and scalable. Starting with a basic pitch template, we randomize features (orientation and viewing position), random placement of objects and backgrounds, and the `visual' characteristics (lighting and weather conditions). We finally render the scene to a database with the associated keypoint data (i.e., the location features within the image).

\paragraph{Player Body/Pose Detection}
There is already a substantial body of work that has focused specifically on the detection and extraction of human poses (describing the bounding area and body keypoints).
For the tests in this paper, we used the PyTorch ResNet101 pre-trained model which produced acceptable results (and could be swapped out for other versions at a later date).
For example, the DeepCut model provides a robust solution for multiple close proximity interactions \cite{pishchulin2016deepcut}.

To ensure the body/pose phase did not detect individuals in the crowd (only providing details for the players), after the football pitch outer marker regions were detected, these were used to `cull' the image to the field area (see Figure \ref{fig:teaser}.

\figuremacroW
{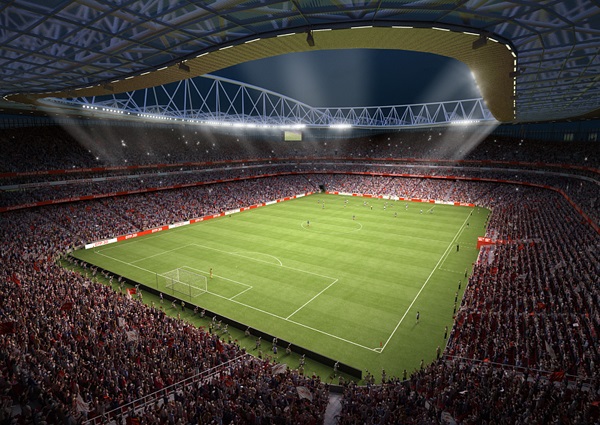}
{Stadium}
{Complexity and the size of the visual can be extended to encapsulate a wide range of visuals, including spotlights, crowds and shadows.}
{1.0}

\section{Results}

Only synthetic (generated) data was used for training.
The generated image datasets where created using web-based tools (i.e., WebGPU API for hardware acceleration \cite{kenwright2022sigwebgpu}). 
We used a web-based solution due to the fact that new technologies offered by modern browsers have greatly increased in capabilities, not to mention, ubiquity across platforms, easy to make iterative updates (during early exploratory stages).
The WebGPU API 
\cite{kenwright2022sigwebgpu} offered GPU-accelerated client-side rendering within the browser (generated 3000 images within a few minutes). %  without plugins.
The training of the network was done offline using PyTorch.
While we experimented with different network configurations initially, the final network topology is shown in Figure \ref{fig:network}.  
This configuration was used for all of the tests and final results.
The web-based resources, generated training data and scripts are accessible online (Git Repository \cite{learningfootballgithub2022}).

\figuremacroF
{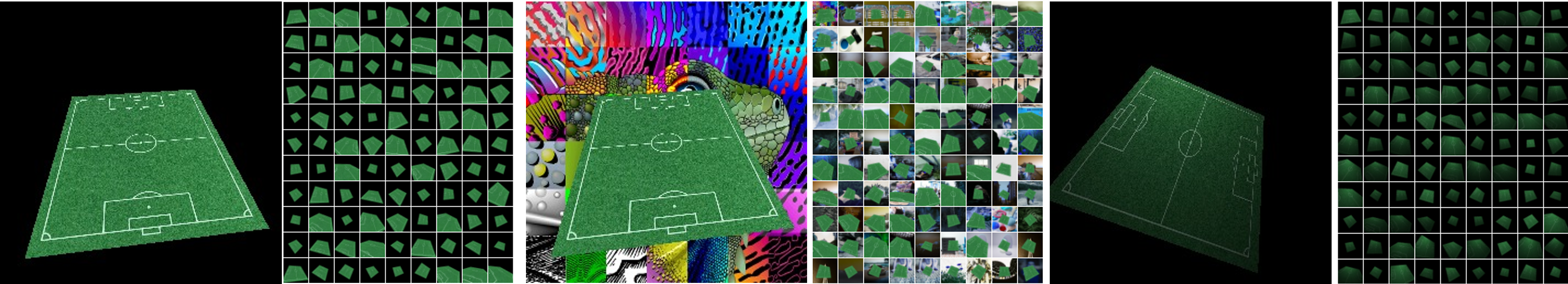}
{Multiple Datasets}
{Created multiple datasets, each dataset consisted of 3000 training images (256x256) with 52 keypoint markers. All images represent 3-dimensional views of a football pitch. (left) basic field with no lighting or background, (middle) random background images and (right) random lighting conditions.
}
{1.0}

\begin{itemize}
\item Synthetic images for training (sample set for training and a corresponding set for validating/checking errors)
\item Simple `clean' image dataset (perfect images with no noise or inconsistencies) - first check to ensure the system is working
\item Complex and diverse dataset (larger range of views) - including `ambiguous' scenes (difficult to extract and identify the pitch data from what the image shows)
\item Mix lighting conditions and noise
\item Add objects and players (ball and players on pitch)

\item Overlay `identified' features on top of 'real-world' data images (compare visually the errors). Such as the location of field markers and the camera frustum.
\item Use different colors to visualize the information (ball, players and the pitch)
\end{itemize}

\figuremacroW
{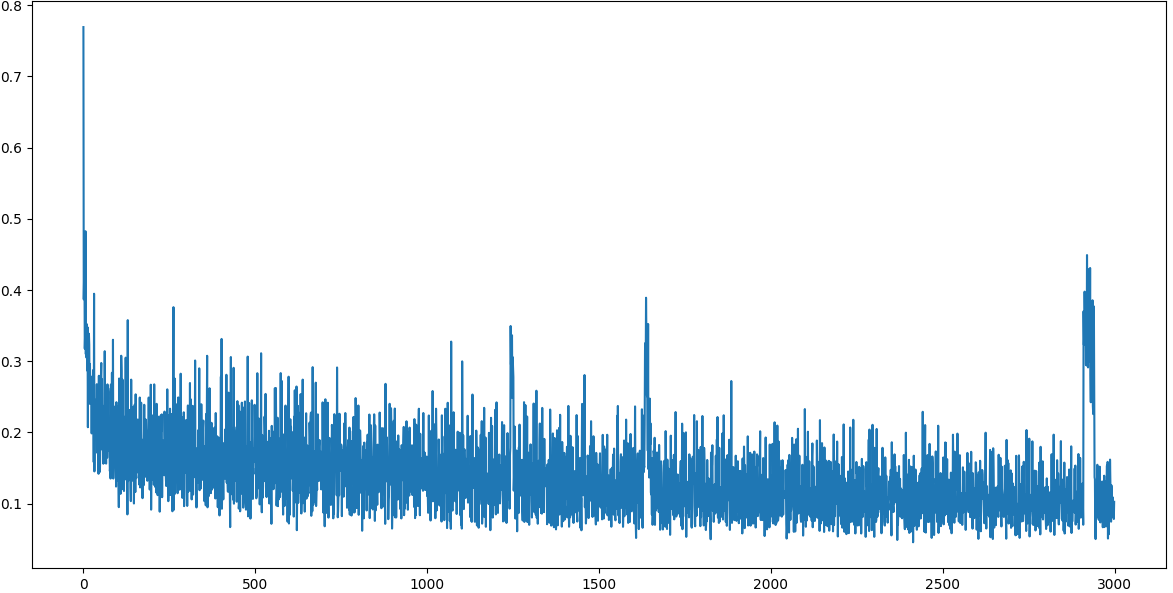}
{Training Plot}
{Fitness of the network as it is trained (500 epoch).}
{0.8}

Evaluating the landmark detectors from the image database, which exhibits environmental, lighting and camera information, we see that models trained with synthetic data alone can generalize well to real data of diverse sources (broadcast media). In order to generate diverse synthetic data, our generative models must be trained with diverse range of configurations.

\figuremacroW
{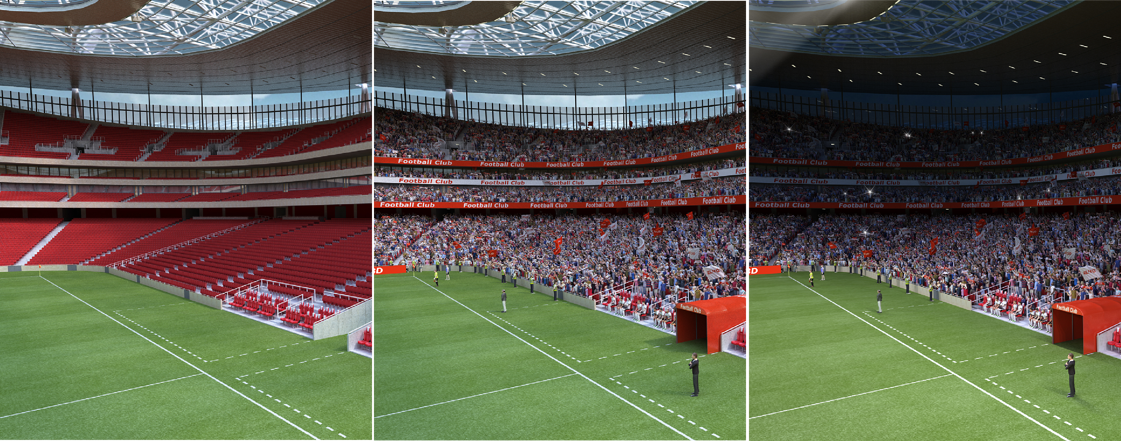}
{Scene and Lighting Conditions}
{Easy to integrate parameters into the generation pipeline to control visual features (e.g., enabling/disabling crowds and lighting conditions). Left middle and right shows the same scene but with no crowd, crowd and amplified darkness). }
{1.0}

\paragraph{Network Architecture}

The final network architecture configuration was reduced to a reasonable size using a combinations of Conv2d and MaxPool2d filter layers before using a series of Dense layers. It required 5 Conv2d $+$ MaxPool2d layers to bring the image dimension down to an acceptable size. 
Dropoff was added between each layer in order to reduce overfitting (increasing the dropoff probability gradually in later layers). Only a single architecture is presented here but other models could be explored for further work (see Figure \ref{fig:network}). 
Training times where generally in the range of half a day with CUDA acceleration (500 epochs).

\paragraph{Number of Epochs and Batch Size}.

Experiment used a batch sizes of 50. We decided to go with batch size of 50 because a larger batch size gave much smoother loss trends over time, which made the loss curve a little easier to interpret, especially for initial experiments where we used a small number of epochs. Also, the computer faced no difficulties in performing the training steps when using batch size of 50.
Initially we ran quick experiments for combinations of different architectures, batch sizes, optimizer and loss functions using 1-5 epochs. For the final setup, we settled on 500 as the training epoch (during testing we found that after 500 epochs the loss decreased very slowly - see Figure \ref{fig:trainingexample}).

\paragraph{Limitations}
The training set is `generated', so care needs to be taken that the `generated' data does not `miss' characteristics that may be visible in real-world data (but not in the generated versions).  Also specific cases where the image does not contain enough information to identify pitch markers (e.g., only grass can be seen).  If the resolution is low (e.g., Figure \ref{fig:pixels}, it can be difficult to identify player/match details - especially when groups of players are in close proximity).

\paragraph{Future Work}
Future work would explore `spatial' models, that is, not just about single frame snap-shots but taking into account multiple frames to extrapolate and extract extra details (e.g., motion analystics and behavioural attributes that maybe difficult or impossible to spot using individual images).

Continue to extend the dataset and provide an online resource (web-based solution) that can be used by anyone to upload and extract features from football images (also video). User could download and use these details for analysis.

Explore fusing multiple dataset resources together to obtain more accurate results, such as, images that record the same football game but from different cameras and viewing locations.
Additional research on improving the keypoint approach, for instance using additional feature locations to improve accuracy \cite{widya2018structure}, also the opportunity to explore the generation and fusion of other data sources (beyond just 2d images) as being explored in related fields \cite{fauzy2022}.

\figuremacroW
{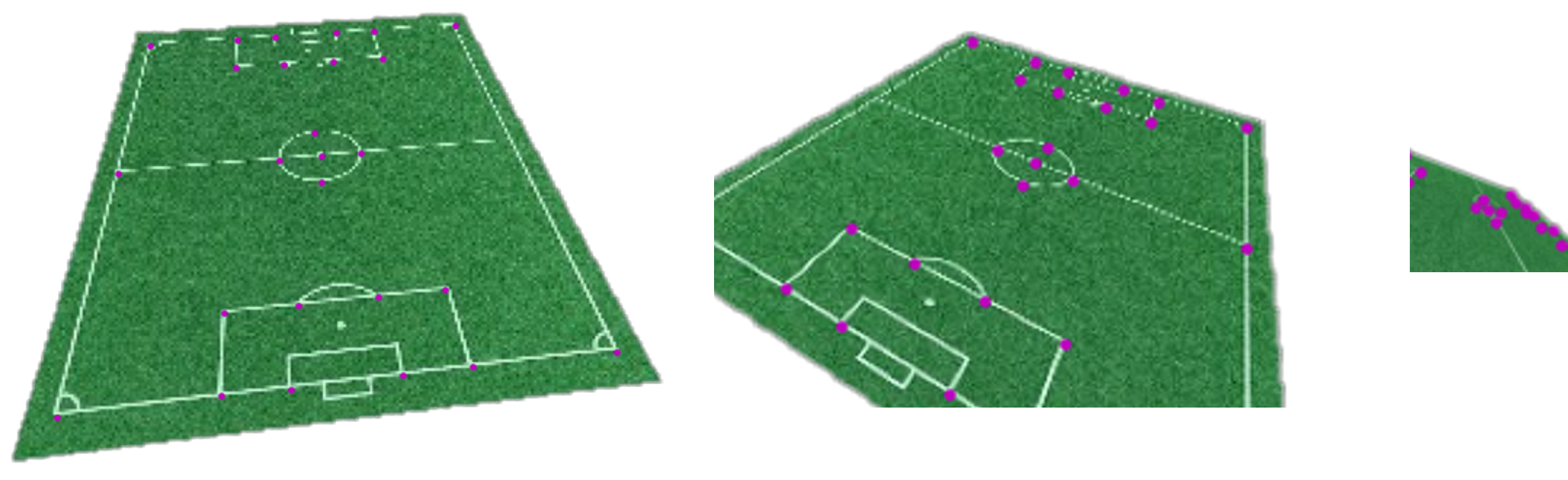}
{Example KeyPoint Data}
{When features are available, even limited views (image on right), the model is able to identify correctly the field points.}
{1.0}

\section{Conclusion/Discussion}

The proposed technique has endless potentials and addresses the lack of automated robust, fast and accurate methods for extracting features from football match images. 

\textbf{Synthetic training data will become more prominent over the years as generated content becomes indistinguishable from the real-world.}  Allows training data to be generated that would otherwise be impossible or rare using real-world data.  While this article has focused specifically on football-games, the concepts are applicable to other areas such as medical operations, vehicle navigation data, flight simulations and even interactive games.

\figuremacroW
{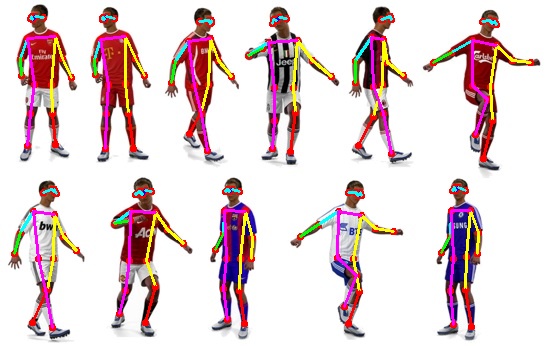}
{Players}
{Characters can be customized and placed in the scene, either to recreate existing games from other data or randomly to create a diverse and complex dataset (includes poses and textures).}
{0.8}

\vspace{-5pt}

\bibliographystyle{ACM-Reference-Format}
\bibliography{paper}

\end{document}